\documentclass[letterpaper, 10 pt, conference]{ieeeconf}  

\IEEEoverridecommandlockouts                              

\usepackage{graphics} 
\usepackage{epsfig} 
\usepackage{mathptmx} 
\usepackage{times} 
\usepackage{amsmath} 
\usepackage{amssymb}  
\usepackage{algorithm}
\usepackage{algorithmic}
\usepackage{pifont}
\usepackage{booktabs}
\usepackage{subcaption}   
\usepackage{multirow}
\usepackage{xcolor}

\title{\LARGE \bf
BFA++: Hierarchical Best-Feature-Aware Token Prune for Multi-View Vision Language Action Model}

\author{Haosheng Li$^{1,2*}$, Weixin Mao$^{3*\dagger}$, Zihan Lan$^{3*}$, 
Hongwei Xiong$^{3}$, Hongan Wang$^{1,2}$,\\ Chenyang Si$^{4}$, Ziwei Liu$^{5}$, Xiaoming Deng$^{1,2\ddagger}$, and Hua Chen$^{3,6\ddagger}$
\thanks{$^{*}$Co-first authors. $^\dagger$Project leader. $^\ddagger$Corresponding authors.}
\thanks{$^{1}$Institute of Software, Chinese Academy of Sciences, Beijing, China.}
\thanks{$^{2}$University of Chinese Academy of Sciences, Beijing, China.}
\thanks{$^{3}$LimX Dynamic, Shenzhen, China.}
\thanks{$^{4}$Nanjing University, Nanjing, China.}
\thanks{$^{5}$Nanyang Technological University, Singapore.}
\thanks{$^{6}$Zhejiang University, Zhejiang, China.}
}

\begin{document}



\maketitle
\begin{abstract}

Vision-Language-Action (VLA) models have achieved significant breakthroughs by leveraging Large Vision Language Models (VLMs) to jointly interpret instructions and visual inputs. However, the substantial increase in visual tokens, particularly from multi-view inputs, poses serious challenges to real-time robotic manipulation. 
Existing acceleration techniques for VLMs, such as token pruning, often result in degraded performance when directly applied to VLA models,
as they overlook the relationships between different views and fail to account for the dynamic and task-specific characteristics of robotic operation.
To address this, we propose BFA++, a dynamic token pruning framework  designed specifically for VLA models. BFA++ introduces a hierarchical pruning strategy guided by two-level importance predictors: an intra-view predictor highlights task-relevant regions within each image to suppress spatial noise, while an inter-view predictor identifies critical camera views throughout different manipulation phases to reduce cross-view redundancy. 
This design enables efficient token selection while preserving essential visual cues, resulting in improved computational efficiency and higher manipulation success rates. 
Evaluations on the RoboTwin benchmark and real-world robotic tasks demonstrate that BFA++ consistently outperforms existing methods. BFA++ improves the success rate by about 10\% on both the $\pi_0$ and RDT models, achieving speedup of 1.8$\times$ and 1.5$\times$, respectively. Our results highlight that context-sensitive and task-aware token pruning serves as a more effective strategy than full visual processing, enabling faster inference and improved manipulation accuracy in real-world robotic systems. 
\end{abstract}

\section{INTRODUCTION}


Recent advances in robotics have been significantly driven by Vision-Language-Action (VLA) models~\cite{rt-2,open-vla,open-vla-oft,tiny-vla,3d-vla,pointvla,efficientvla,Mobility-vla,pi0,robomatrix,Bi-vla,hbridvla,Autoregressive_action_sequence}, which integrate visual perception and natural language understanding to enable robots to perform complex tasks from human instructions. 
To further enhance perceptual capability and action precision, a notable trend in recent research is the use of multi-view visual inputs. This approach provides richer visual observations by capturing the scene from multiple camera views and is especially beneficial for dual-arm systems that require fine-grained coordination between manipulators, such as RDT~\cite{rdt} and $\pi_0$~\cite{pi0}. 

However, while multi-view settings provide richer information, they also bring challenges for VLA models. Unlike VLM domains with abundant training data, robotics datasets are inherently limited, resulting in VLA models having less well-established attention mechanisms. Consequently, these models struggle to distinguish task-relevant visual cues from redundant information across multiple views. Without proper guidance, this redundancy can misdirect attention, causing critical manipulation cues to be overlooked. This challenge makes incorporating human manipulation priors into VLA both practical and necessary~\cite{bfa,efficientvla,SP-VLA}.

\begin{figure}[t]
  \centering
    \includegraphics[width=\linewidth]{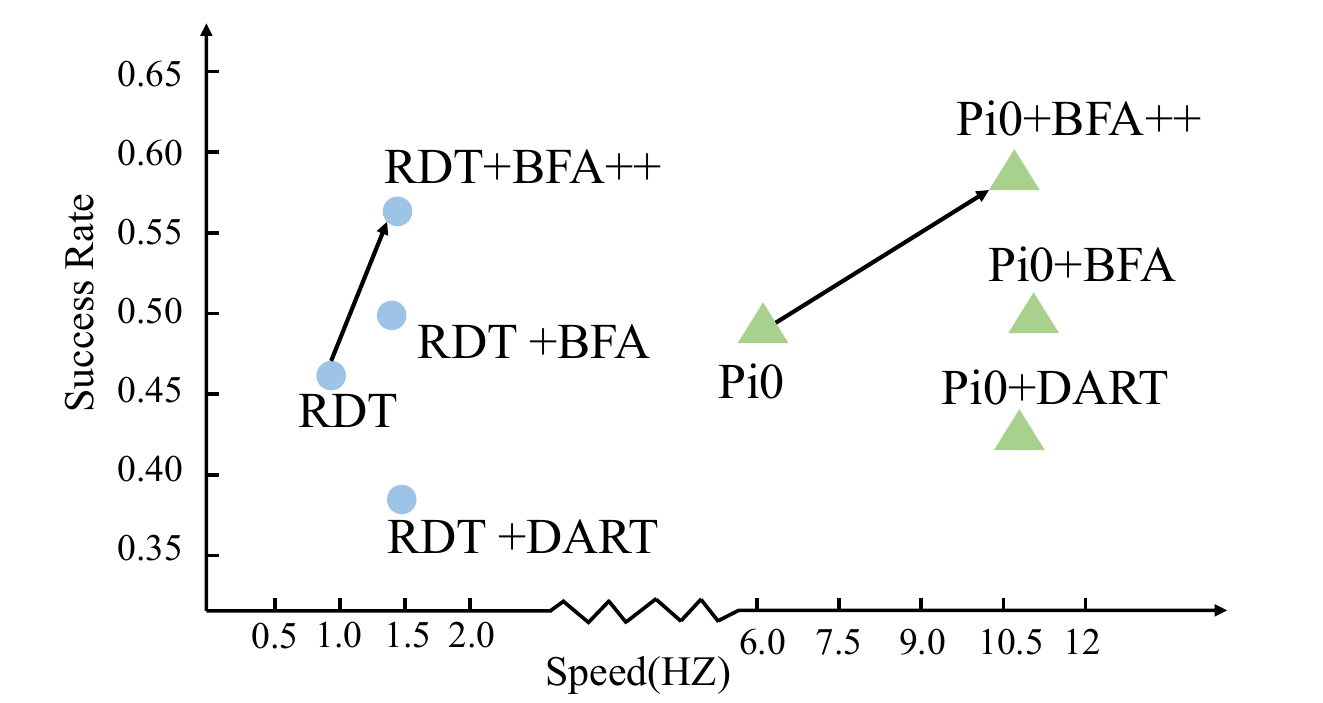}
    
   \caption{The comparison between BFA++ and other methods. Compared to existing token pruning methods, our approach not only improves inference speed but also significantly increases the success rate.}
   \label{overview_plot}
   \vspace{-0.5cm}
\end{figure}

In VLA models, limited pretraining data and fixed action output spaces constrain the model's ability to learn robust attention mechanisms~\cite{tiny-vla,simplevla,InSpire}. This setting gives rise to two major challenges in attention allocation. First, at the inter-view level, standard VLA training cannot reliably estimate the relative importance of different camera views, which changes dynamically depending on the task phase. Second, at the intra-view level, models may focus on irrelevant regions, such as backgrounds or disrupting objects, rather than task-aware areas. Although using unsupervised token importance predictors~\cite{dynamicvit} and general unsupervised pruning methods~\cite{efficientvla,SP-VLA,DART,fastV} show promise in other domains like VLMs and image classification, they fail in VLA scenarios due to the lack of proper supervisory signals to identify task-relevant regions and ignore the relationships between multiple images. Without supervision, these methods tend to treat all tokens equally and can even degrade performance, as the long optimization chain from action head to LLM to image encoder makes the supervision signals unclear, highlighting the need for direct guidance in manipulation scenarios. 

To address these attention allocation challenges, we conducted qualitative experiments detailed in Sec~\ref{sec:Inter-View_and_Intra-view_Importance}, we find that inter-view importance varies dynamically: the head view suffices for approach and post-manipulation phases, while the wrist view is crucial during manipulation. Within each view, attention should focus on task-relevant regions such as the end-effector and target objects, while avoiding background distractions.

Based on the above challenges and findings about two-level importance variations, we propose a supervised dynamic token pruning method specifically designed for VLA post-training. It employs a hierarchical local-global token pruning strategy: first, locally pruning less important tokens within each view; then, globally ranking all remaining tokens by their combined importance scores to eliminate the least critical tokens across all views. We design two networks for hierarchical token pruning: an intra-view importance predictor (Intra-IP) identifies crucial tokens (e.g., robotic grippers, target objects) within each view, and an inter-view importance predictor (Inter-IP) determines critical views across manipulation stages. The training data is annotated using our comprehensive annotation system. Inter-view importance can be annotated through multiple approaches: LLM-based annotation~\cite{Gemini}, bounding box detection~\cite{Florence-2,sam2} with hard-coded rules, or human annotation. Meanwhile, intra-view importance is annotated using task-oriented bounding box prediction methods~\cite{Grounding_dino,Florence-2,sam2}. Both predictors are jointly trained with the backbone during post-training, enabling robust elimination of redundant tokens while preserving task-critical information.

Our method produces better-distributed visual tokens with reduced similarity, enabling more accurate manipulation (Figure~\ref{overview_plot}). Experiments on RoboTwin~\cite{robotwin} and real environments demonstrated that our method achieves 1.8$\times$ speedup with 10\% success improvement on $\pi_0$~\cite{pi0} and 1.5$\times$ speedup with 10\% improvement on RDT~\cite{rdt}.
Our main contributions can be summarized as follows: 1)  We design a hierarchical token pruning method using inter-view and intra-view importance scores for VLA models to robustly eliminate redundant tokens while preserving task-aware information. 
2) Experiments demonstrate that our plug-and-play pruning framework can be used effectively in both $\pi_0$~\cite{pi0} and RDT~\cite{rdt} models, which can achieve 1.5-1.8× speedup with improved about 10\% success rates.

\section{Related Works}
\subsection{Vision Language Action Model}

Vision-Language-Action (VLA) models have emerged as a promising paradigm for robotic manipulation, integrating visual perception, language understanding, and action generation. Early VLA models primarily used single-image inputs. Among these, RT-2~\cite{rt-2} were pioneering works that demonstrated large-scale data collection and transformer architectures could enable robust real-world robotic control, and that web-scale pre-training effectively transfers to embodied tasks. OpenVLA~\cite{open-vla} proposed an open-source VLA trained on diverse robot demonstrations with strong generalization. TinyVLA~\cite{tiny-vla} explored model compression for resource-constrained deployment.

Researchers later recognized that single images provide insufficient information, leading to multi-image VLA approaches. $\pi_0$~\cite{pi0} and $\pi_{0.5}$~\cite{pi0.5} introduced flow matching techniques and take three images as input. RDT~\cite{rdt} proposed a unified framework integrating multiple images for enhanced spatial-temporal understanding. OpenVLA-OFT~\cite{open-vla-oft} extended OpenVLA~\cite{open-vla} with dedicated action heads and multi-image inputs to improve performance on complex tasks.
Other works explored additional modalities, such as 3D-VLA~\cite{3d-vla} with 3D scene understanding and PointVLA~\cite{pointvla} with point cloud representations.

\begin{figure*}[t]
  \centering
    \includegraphics[width=\linewidth]{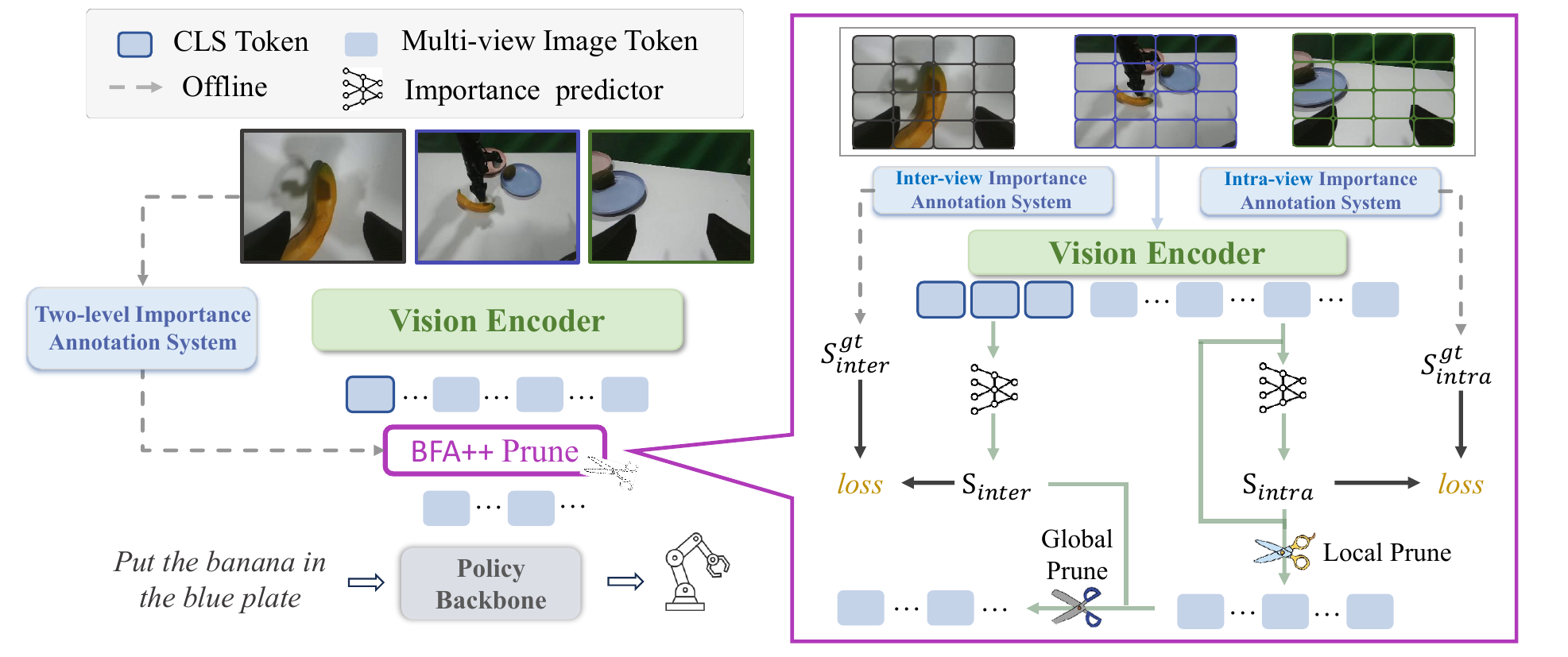}
    
   \caption{The pipeline of BFA++. The left figure shows our plug-and-play module for token pruning, and the right figure further details it. After obtaining the offline annotated inter-view importance $S^{gt}_{inter}$ and intra-view importance $S^{gt}_{intra}$, we train the two importance predictor, and predict these two importance scores ($S_{inter}$, $S_{intra}$). Based on the predicted intra-view importance, we perform local pruning on the tokens. Then, based on both inter-view importance and intra-view importance, we conduct global pruning to obtain the final pruned tokens.}
   \label{main_pipeline}
   \vspace{-0.5cm}
\end{figure*}

\subsection{Token Prune Method}
Token pruning has emerged as an effective technique to reduce computational costs in vision transformers by dynamically removing redundant tokens during inference~\cite{bfa,dynamicvit,efficientvla,fastV,token_merge,DART}. DynamicViT~\cite{dynamicvit} pioneered this approach by introducing a learnable prediction module that identifies and prunes less important tokens at different transformer layers. Token Merging~\cite{token_merge} proposed merging similar tokens rather than discarding them, using bipartite soft matching to preserve information while reducing the token count.
Recent advances have improved pruning strategies. But most of them primarily focus on pruning within transformer backbones based on token similarity and attention scores. FastV~\cite{fastV} used attention scores to identify important tokens with minimal changes to the architecture. DART~\cite{DART} introduced token ranking by computing visual token similarity. BFA~\cite{bfa} developed a budget-aware framework that adjusts fusion ratios based on inter-view importance. EfficientVLA~\cite{efficientvla} and SP-VLA~\cite{SP-VLA} use attention-based or spatial heuristics but lack task-specific token selection guidance. However, different from our  BFA++, nearly all these methods were primarily designed for general vision-language models, and these methods~\cite{efficientvla,SP-VLA,DART,fastV} fail to dynamically adapt token pruning to specific manipulation requirements, potentially discarding action-critical visual information. Our BFA++ addresses this issue by task-aware dynamic token selection tailored to robotic manipulation.

\section{Method}
We propose BFA++ (Figure~\ref{main_pipeline}), a hierarchical token pruning framework for VLA models that accelerates inference and improves task success rates by filtering visual information. We use the inter-view importance score to evaluate the importance of each view, and use intra-view importance to rate the importance of tokens within view. Then, we locally prune less important tokens within each view using the intra-view importance score, and  globally rank all remaining tokens by intra-view and inter-view importance scores to eliminate less critical tokens across all views.

We first analyzed the inter-view and intra-view importance during the manipulation process in Sec.~\ref{sec:Inter-View_and_Intra-view_Importance}.
Then we detail our offline preparation of supervisory data for two-level importance in Sec.~\ref{sec:Offline_Data_Annotation}. Finally, we present the BFA++ pruning pipeline and its integration into VLA post-training in Sec.~\ref{sec:bfa_pipeline}. 

\subsection{Intra-View and Inter-View Importance of Token During Manipulation}
\label{sec:Inter-View_and_Intra-view_Importance}
\begin{figure}[h]
  \centering
    \includegraphics[width=\linewidth]{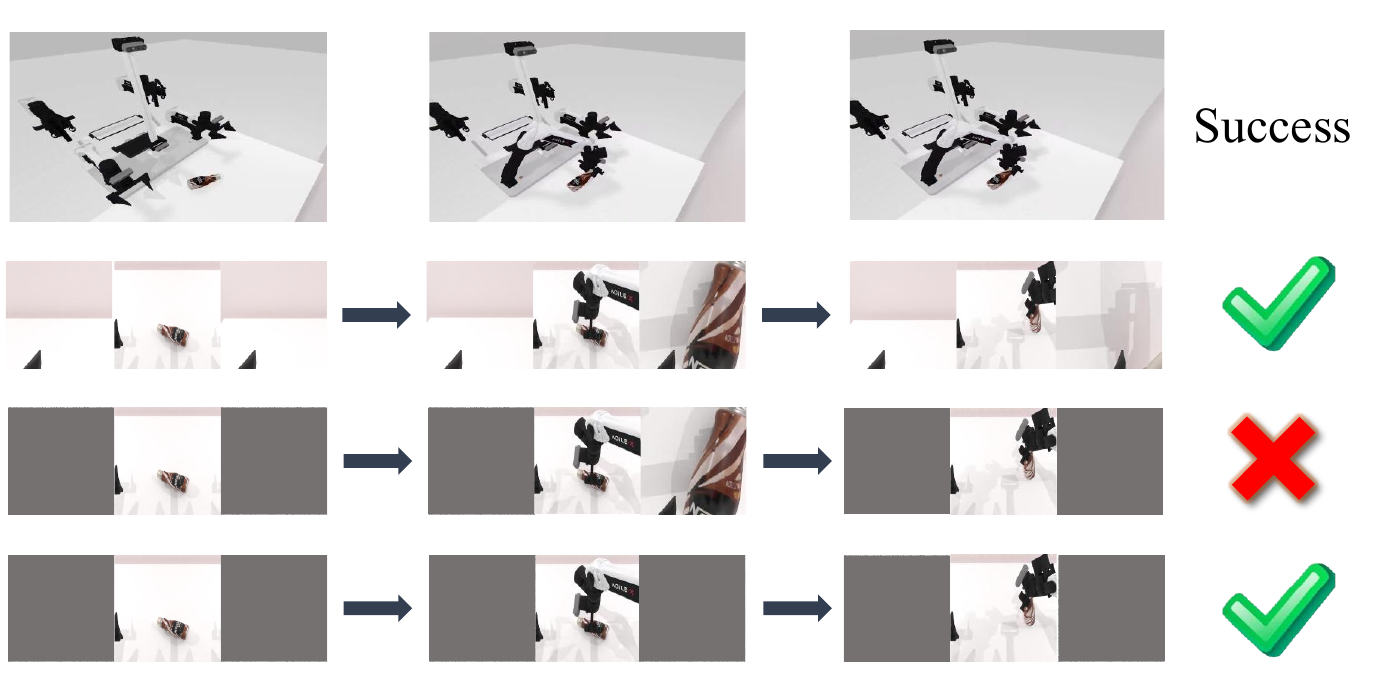}
    
   \caption{Analyzing the importance of different viewpoints. We remove one or two view images to see how the performance of a trained $\pi_0$ model is affected. This experiment demonstrates that the importance of different viewpoints varies across different stages of operation.} 
   \label{View_importance}
   \vspace{-0.3cm}
\end{figure}

Understanding the importance of different tokens across and within camera views is crucial for effective multi-view robotic manipulation. Within each view, models should focus on task-oriented regions, such as the robotic end-effector and target objects, rather than scattering attention across background elements or irrelevant scene components, as also demonstrated by methods~\cite{3D_CAVLA,InSpire}.

Moreover, to demonstrate the dynamic nature of token importance, we conduct perturbation experiments by excluding different camera views across manipulation stages, as shown in Figure~\ref{View_importance}. The experimental results reveal significant variations in the impact of view removal across the manipulation phases. When the wrist camera input is occluded during the approach or post-manipulation stages, it does not affect the success rate. However, during the manipulation stage, occluding the wrist camera causes the experiment to fail.
These findings indicate that camera importance varies dynamically, consistent with BFA~\cite{bfa}: the head view suffices during approach phases, whereas the wrist camera becomes critical during fine manipulation when the head view is occluded.

\subsection{Offline Two-Level Importance Annotation}
\label{sec:Offline_Data_Annotation}

\begin{figure}[h]
  \centering
    \includegraphics[width=\linewidth]{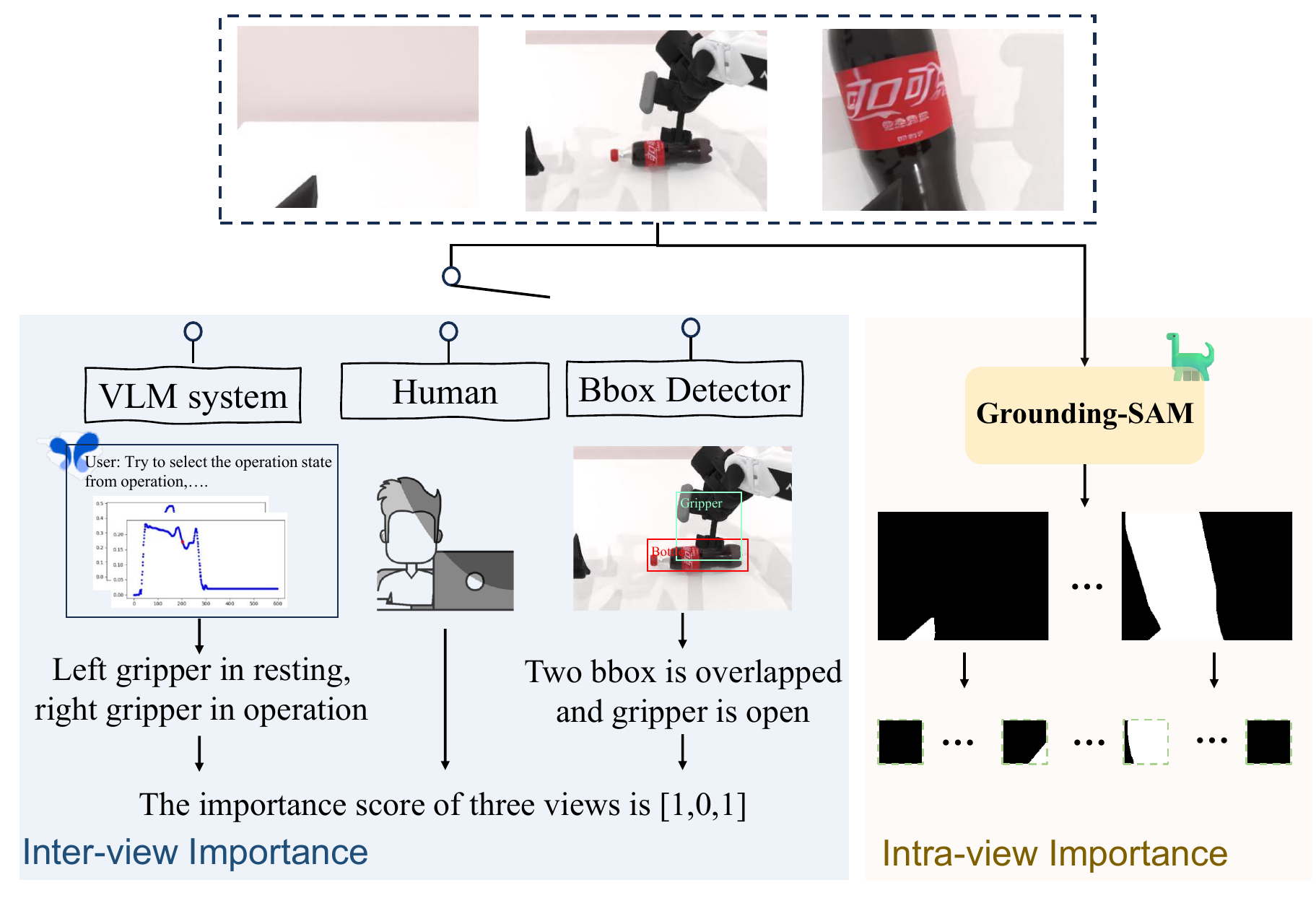}
    
   \caption{The annotation system of BFA++. For inter-view importance, we provide three optional methods (VLM, manual annotation, bounding boxes to detect overlaps). For intra-view importance, we use the Grounding-SAM method to identify task-related regions. Details of the VLM annotation method can be found in BFA~\cite{bfa}.}
   \label{Anno_sys}
   \vspace{-0.3cm}
\end{figure}
Based on the findings above, we built a two-level offline importance annotation system to separately annotate the inter-view importance and the intra-view importance across tokens within each view, providing explicit supervision during post-training. 

For intra-view token importance, we use off-the-shelf models to select task-relevant regions. Typical approaches include background removal or bounding box detection method~\cite{grounding_dino1.5,Grounded_sam,Florence-2} to identify task-relevant areas. We selected the bounding box detection method for annotation because of its efficiency and suitability for our specific task requirements.

For inter-view importance, we set the importance score to 1 for cameras that capture detailed interactions: specifically, when a manipulator interacts with objects, both the corresponding wrist and head cameras receive an importance score of 1. At all other times, only the head camera has an importance score of 1, while both wrist cameras are set to 0. 
We use three alternative methods for annotating inter-view importance. First, following BFA~\cite{bfa}, we use large language models~\cite{Gemini} that analyze gripper trajectories, rest pose data, and gripper open/close status to determine when each manipulator is actively interacting with objects. Second, we employ a bounding box detector~\cite{grounding_dino1.5,Florence-2} to detect overlaps between object and gripper bounding boxes, identifying interaction moments through visual detection. Third, we offer manual annotation as an option, where experienced annotators can complete one task in approximately one hour. These three methods categorize the manipulator states of each arm into phases (approaching, starting operation, moving with object, and retracting) and combine the states of both arms to determine which cameras should be marked as important at each timestamp.

With the generated mask, we map it back to image patches to obtain importance scores for each patch (token). If a patch contains pixels within the mask, we assign it a score of 1, otherwise, 0.

\subsection{BFA++ Token Prune}
\label{sec:bfa_pipeline}

To address redundant input information in VLA models, we propose BFA++, a token pruning method that dynamically prunes redundant tokens during operation by predicting each token's importance.
Our BFA++ framework consists of three key components: 1) Two-level importance predictors, 2) a hierarchical token pruning strategy, and 3) integration with VLA post-training. 

\vspace{1mm}
\noindent \textbf{Two-level Importance Predictors.} As shown in  Figure~\ref{main_pipeline}, we design two lightweight neural networks with the same architecture to predict importance scores at different levels. The inter-view importance predictor $f_{\text{inter}}$ takes CLS tokens $\{T_{cls}^{v}\}_{v=1}^V$ of all views as input (in our experiment, $V=3$), which is the output of the vision encoder from all views, and outputs importance weights $S_{inter}=[S_{inter}^{1},S_{inter}^{2},S_{inter}^{3}]
$ for each view:
\begin{equation}
S_{inter} =  f_{\text{inter}}([T_{cls}^{1},T_{cls}^{2},T_{cls}^{3}]).
\end{equation}

The intra-view importance predictor $f_{\text{intra}}$ takes individual visual token $T_{\text{Image}}^{v,n}$ from the vision encoder as input, and outputs intra-view importance score $S_{intra}^{v,n}$ of the $n$-th token of the $v$-th image:
\begin{equation}
S_{intra}^{v,n}=  f_{\text{intra}}(T_{Image}^{v,n})
\end{equation}

In this way, we obtain two-level token importance through these two lightweight networks.
    
\begin{figure}[th]
  \centering
    \includegraphics[width=\linewidth]{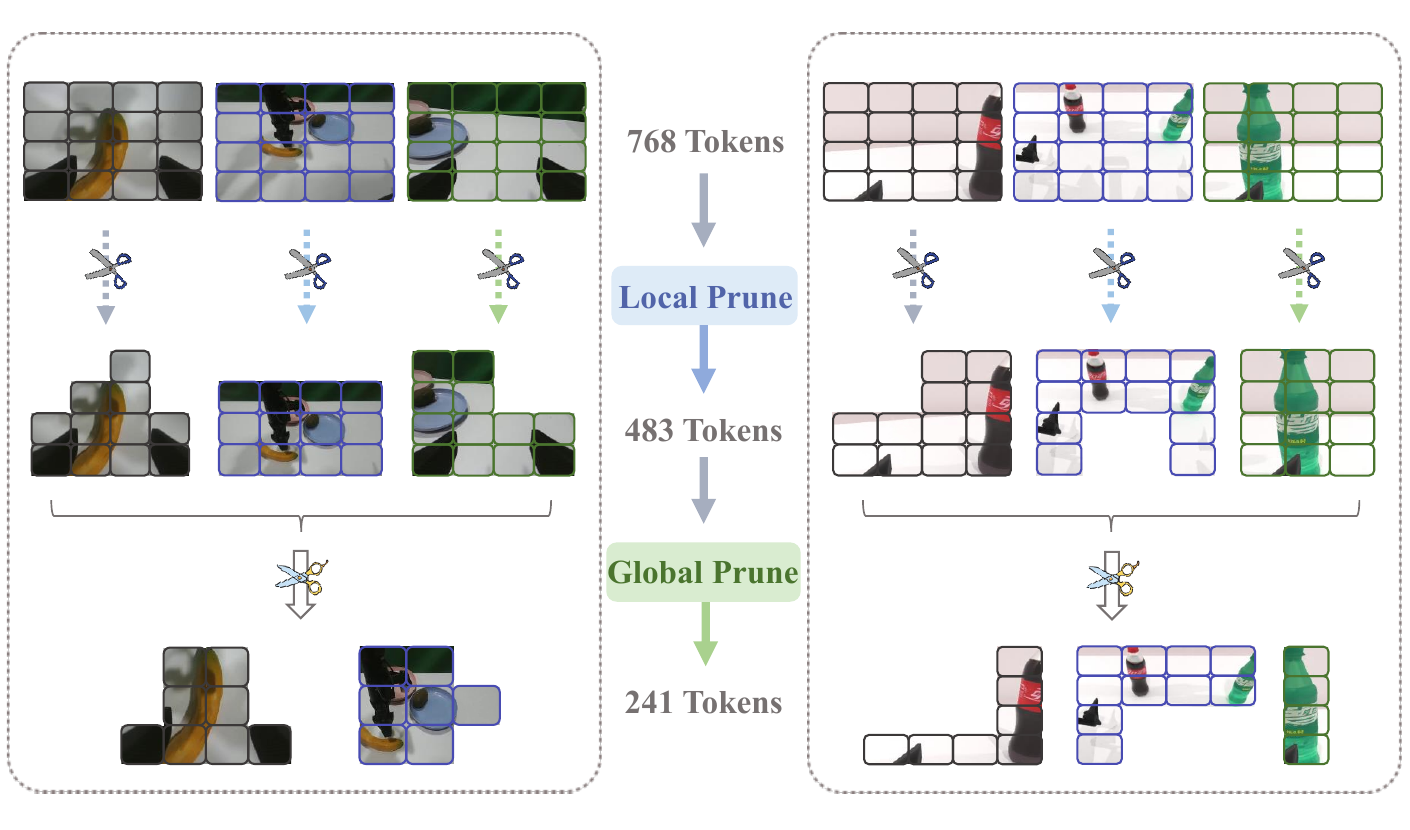}
    
   \caption{The local and global token prune in $\pi_0$+BFA++.}
   \label{detail_prune}
   \vspace{-0.3cm}
\end{figure}

\vspace{1mm}
\noindent \textbf{Hierarchical Token Pruning.} Given the predicted two-level importance scores $S_{inter}$ and $S_{intra}^{v,n}$, we perform token pruning in two stages: local pruning and global pruning, shown in Figure~\ref{detail_prune}. 

\vspace{1mm}
\noindent \textit{Local Pruning:} We use the intra-view importance scores to perform local pruning within each view. At this stage, we rank tokens within each view by their intra-view importance scores and remove a fixed proportion of the least important tokens. However, to ensure that our score map is spatially coherent and avoid abrupt changes in importance values, we apply spatial adaptive weighting (Eq.~\ref{eq-saw}) to refine the raw importance scores before pruning. Specifically, each token's final importance score is refined by incorporating information from spatially neighboring tokens, where closer tokens have a stronger influence on the final score: 
\begin{equation} 
S_{intra}^{v,n}= \sum_{i=1}^{N}{\frac{S_{intra}^{v,i}}{d(pos_n, pos_i)+\epsilon}}
\label{eq-saw}
\end{equation} 
where $N$ is the number of tokens output by the vision encoder for each image,  $d(pos_n, pos_i)$ represents the spatial distance between the 2D positions of tokens $n$ and $i$ in the feature map, and $\epsilon$ is a small constant to prevent division by zero. This adaptive weighting creates spatially smooth importance distributions and avoids isolated high-importance regions. 
We then normalize the scores within each view and independently prune the $\alpha_v$ ($v \in \{1,2,3\}$) ratio of tokens with the lowest intra-view importance scores for each image view.


\textit{Global Pruning:} After local pruning, we perform global pruning to further reduce tokens by ranking all remaining tokens in all views and removing less important tokens.

We fuse inter-view and intra-view importance by multiplying each token's importance score with its view weight:
\begin{equation} 
S_{final}^{v,n} = S_{inter}^{v}  \cdot S_{intra}^{v,n} 
\end{equation}
where $S_{final}^{v,n}$ is the final global importance score for token $n$ in view $v$, and $S_{inter}^{v}$ is the importance weight for view $v$.

Next, we rank remaining tokens by their final global importance scores $S_{final}^{v,n}$. This ranking considers both how important each token is within its view and how important that view is with respect to all views.

Finally, we remove the tokens with the lowest $\beta$ proportion from the final scores in this unified ranking. This ensures that we keep the most important tokens regardless of which view they come from.


\vspace{1mm} 
\noindent \textbf{Integration with VLA Post-Training.} During VLA post-training, we jointly optimize the importance predictors Inter-IP and Intra-IP alongside the main model. Based on the standard action prediction loss $L_{\text{action}}$ of the baseline VLA model, we introduce two auxiliary losses.
The inter-view importance loss $L_{\text{inter}}$ supervises inter-view importance:
$L_{\text{inter}} = \text{BCE}(f_{\text{inter}}([T_{cls}^{1},T_{cls}^{2},T_{cls}^{3}]), S_{\text{inter}}^{\text{gt}})$
where $f_{\text{inter}}$ is the inter-view importance predictor that takes concatenated CLS tokens from three views as input.
The intra-view importance loss $L_{\text{intra}}$ guides intra-view importance based on task-oriented masks:
$L_{\text{intra}} = \sum_{v=1}^{V}\sum_{n=1}^{N} \text{BCE}(S_{\text{intra}}^{v,n}, S_{\text{intra}}^{v,n,\text{gt}})$
where $S_{\text{intra}}^{v,n,\text{gt}}$ and $S_{\text{inter}}^{\text{gt}}$ denote the ground truth intra-view and inter-view importance annotated by our annotation system. Both auxiliary losses use binary cross-entropy (BCE) loss. The total training loss is formulated as:
\begin{equation}
{L}_{\text{total}} = {L}_{\text{action}} + \lambda_1 {L}_{\text{inter}} + \lambda_2 {L}_{\text{intra}}
\end{equation}
where $\lambda_1, \lambda_2$ are two weighting coefficients. In our experiments, we find that both two-level predictors converge quickly, and we can use the same hyperparameters as the original baseline for all other parameters.

\section{Experiment}

\subsection{Implementation Details}
\begin{figure}[h]
  \centering
    \includegraphics[width=\linewidth]{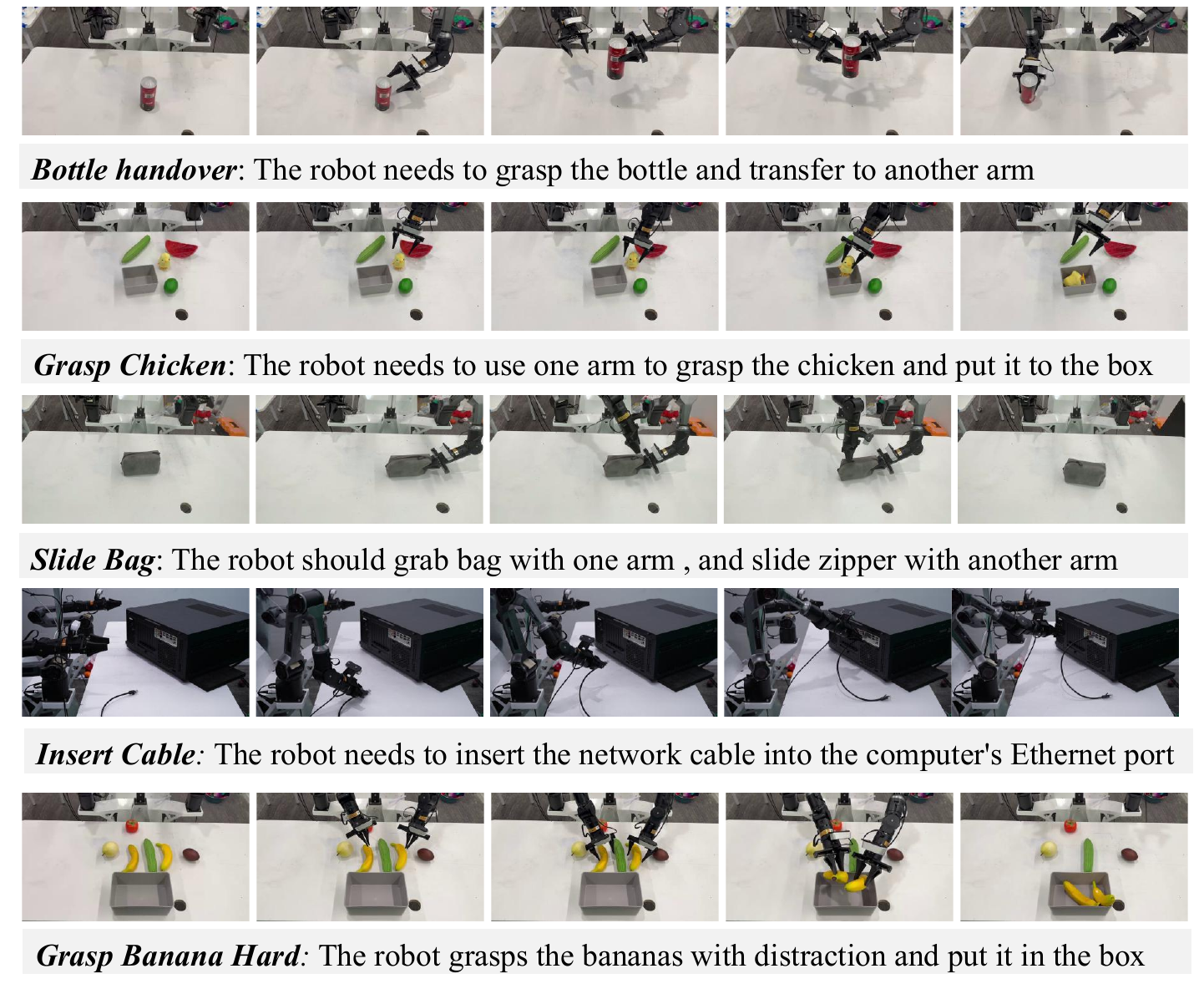}
    
   \caption{Task description of real world experiments.}
   \label{Task_description}
   \vspace{-0.5cm}
\end{figure}

We evaluate our algorithm in both RoboTwin and real-world environments. During comparison, all baseline methods and our approach are trained on the same dataset for the same number of steps.  We use the officially released codes of $\pi_0$~\cite{pi0} and RDT-1B~\cite{rdt}.
For $\pi_0$~\cite{pi0} and algorithms built on $\pi_0$~\cite{pi0}, we perform full-parameter fine-tuning for 5000 steps with a global batch size of 256 for each task. For RDT~\cite{rdt} and algorithms built on RDT, we fine-tune for 1200 steps with a global batch size of 128. The loss hyperparameters $\lambda_1$ and $\lambda_2$ are both set to 0.1, the prune ratio $[\alpha_1,\alpha_2,\alpha_3,\beta]$ set to be $0.3,0.2,0.2,0.5$ and the weighting parameter $\epsilon$ is set to 0.01.
Since dropping tokens within the LLM backbone in $\pi_0$~\cite{pi0} would break the KV cache which will slow the speed, we perform token pruning before tokens enter the LLM backbone. For RDT~\cite{rdt}, following DART~\cite{DART}, we perform token pruning at the second layer of the DiT block. Other methods maintain complete consistency with their original papers.
We collected 50 episodes for the 'Bottle Pick' task and 100 episodes for the seven other simulator tasks. For the four out-of-domain (OOD) task in RoboTwin2~\cite{robotwin2}, we collect 50 episodes each. Additionally, we designed 5 real-world experiments, each with 200 episodes, encompassing single-arm and dual-arm tasks as shown in Figure~\ref{Task_description}, with several tasks featuring extensive distractions that distinguish them from simulator environments. For evaluation, we test each simulator task 100 times and each real-world task 20 times. All training was conducted on eight NVIDIA A100 GPUs, and inference was performed on a single NVIDIA RTX 3090 GPU.

\subsection{Experiment Results}

\begin{table*}[htbp]
\setlength{\tabcolsep}{2pt}
\centering
\caption{Comparison of baseline and BFA++ in RoboTwin. Our method surpasses nearly all baseline methods across various metrics.}
\vspace{-0.1cm}
\label{tab:model_performance}
\renewcommand{\arraystretch}{1.2}
\begin{tabular}{c | ccccc ccc | c | c}
\hline
\textbf{Model} & \textbf{Hammer Beat} & \textbf{Block Handover} & \textbf{Bottles Pick} & \textbf{Blocks Stack} & \textbf{Mug Hang} & \textbf{Bottle Adjust} & \textbf{Shoe Place} & \textbf{Cup Place} & \textbf{Mean} & \textbf{FPS} \\
\hline
$\pi_0$~\cite{pi0} & 0.78 & 0.52 & 0.65 & 0.54 & 0.07 & 0.46 & 0.59 & 0.36 & 0.496 & 6.5 Hz \\
$\pi_0$+DART~\cite{DART} & 0.70 & 0.53 & 0.64 & 0.41 & 0.08 & 0.50 & 0.49 & 0.30 & 0.456 & 9.8 Hz \\
$\pi_0$+BFA~\cite{bfa} & 0.80 & 0.62 & 0.59 & 0.60 & 0.07 & 0.52 & \textbf{0.65} & 0.34 & 0.524 & 11.1 Hz \\
$\pi_0$+BFA++ & \textbf{0.84} & \textbf{0.66} & \textbf{0.70} & \textbf{0.60} & \textbf{0.15} & \textbf{0.58} & \textbf{0.65} & \textbf{0.48} & \textbf{0.583} & 10.3 Hz \\
\hline
RDT~\cite{rdt} & 0.83 & 0.78 & 0.48 & 0.15 & 0.05 & 0.48 & 0.61 & 0.38 & 0.470 & 1.0 Hz \\
RDT+DART~\cite{DART} & 0.71 & 0.70 & 0.58 & 0.07 & 0.03 & 0.48 & 0.41 & 0.24 & 0.403 & 1.6 Hz \\
RDT+BFA~\cite{bfa} & 0.84 & 0.80 & 0.59 & \textbf{0.22} & 0.05 & 0.42 & 0.59 & 0.40 & 0.489 & 1.7 Hz \\
RDT+BFA++ & \textbf{0.89} & \textbf{0.82} & \textbf{0.80} & 0.19 & \textbf{0.09} & \textbf{0.58} & \textbf{0.67} & \textbf{0.48} & \textbf{0.565} & 1.5 Hz \\
\hline

\end{tabular}
\vspace{-0.5cm}
\end{table*}
We evaluate the effectiveness of our algorithm in both simulation and real-world environments, and the results of success rate and inference speed in control FPS are shown in Table~\ref{real_world_result} and Table~\ref{tab:model_performance}.
We select two state-of-the-art methods $\pi_0$~\cite{pi0} and RDT~\cite{rdt} as VLA baseline models, and BFA~\cite{bfa} and one of the best VLM token pruning methods--DART~\cite{DART} for  comparison of pruning. All methods are retrained during post-training.

\begin{figure*}[t]
  \centering
\begin{tabular}{c}
\includegraphics[width=0.45\linewidth]{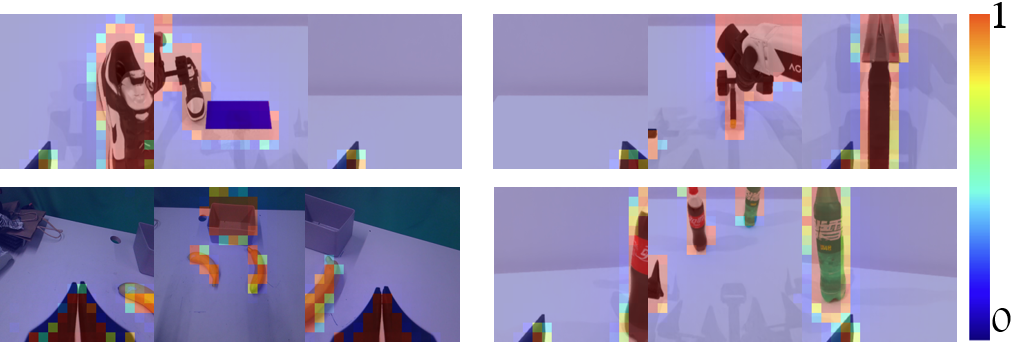} \\
(a) Intra-view importance visualization \\
\includegraphics[width=0.45\linewidth]{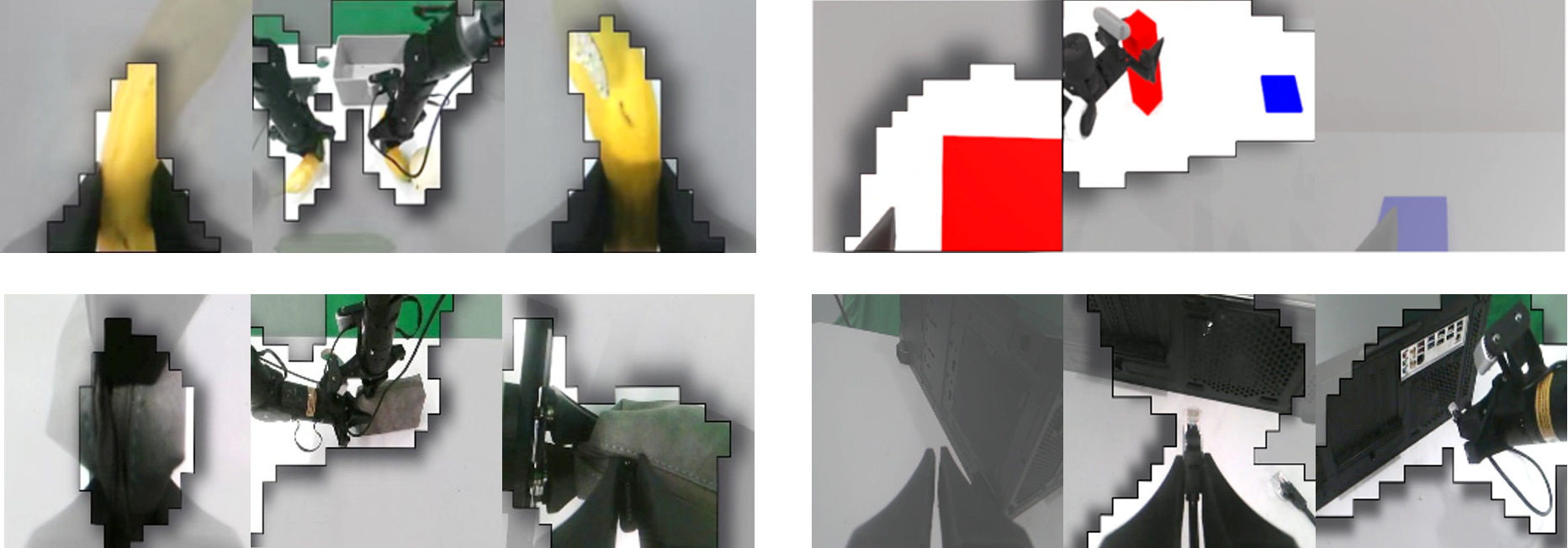}\\
(b) Final prune visualization
\end{tabular}
\begin{tabular}{c}
\includegraphics[width=0.4\linewidth]{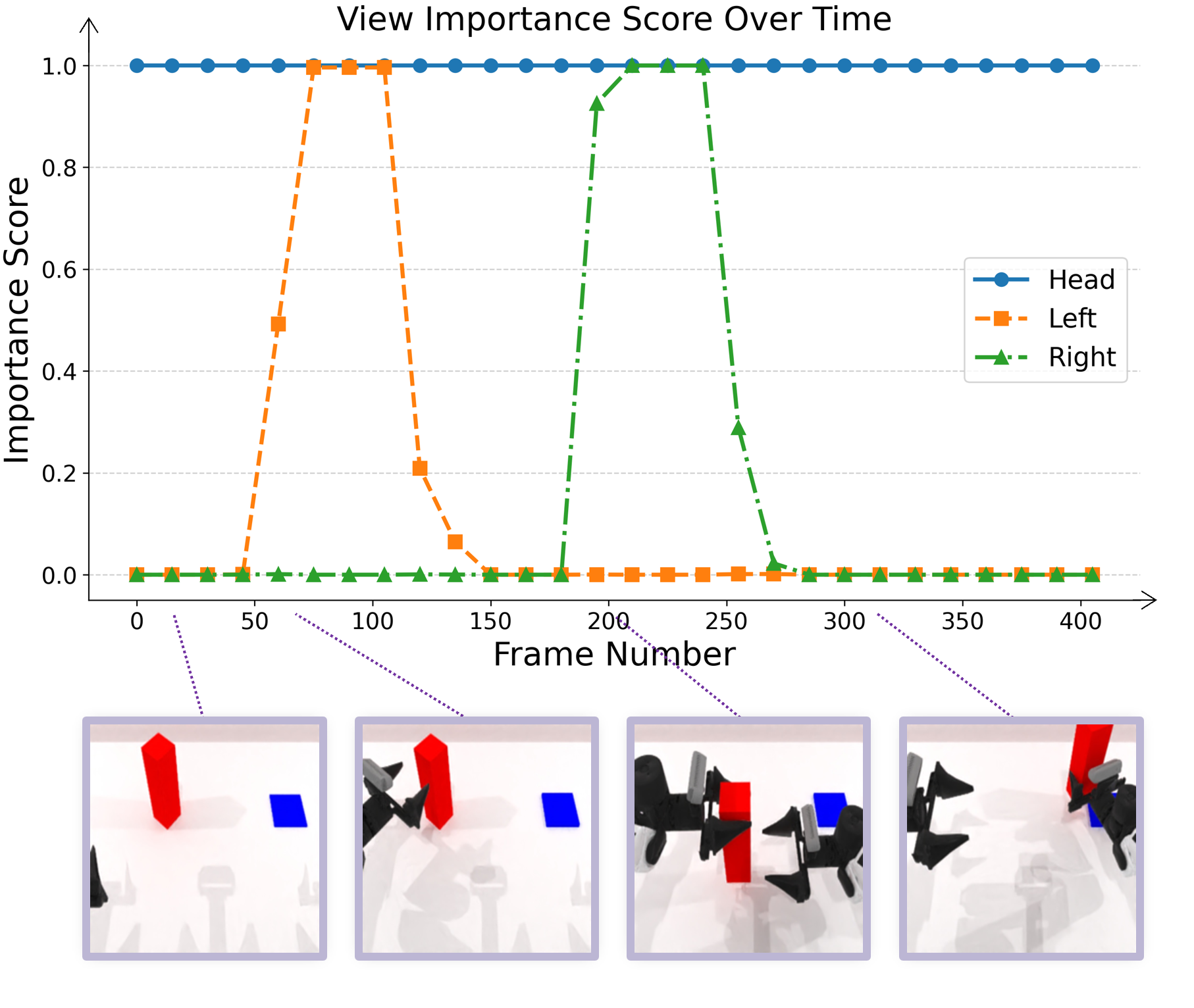}\\
(c) Inter-View importance visualization
\end{tabular}
   \caption{The visualization of the final pruned tokens with BFA++, predicted intra-view importance, and inter-view importance. In figures (a) and (b), each image is composed of three views (left, head, right wrist). Figure (a) represents the token importance predicted by Intra-IP, while figure (b) shows the final token retention visualization. Figure (c) illustrates how the inter-view importance predicted by Inter-IP changes throughout the manipulation process shown in the bottom portion.}
   \label{process}
   \vspace{-0.3cm}
\end{figure*}

Our method achieves significantly higher success rates and faster control frequencies than baselines, with only a slight reduction in speed compared to BFA~\cite{bfa} due to the additional computation of intra-view importance scores alongside inter-view scores.
We observe that the DART~\cite{DART} pruning method tends to perform chaotic and disorganized pruning, preferring to keep only the middle portions while discarding the gripper's spatial position information. Our method does not suffer from this issue. Moreover, to demonstrate the effectiveness of our method on OOD tasks, we conducted tests on the RoboTwin2~\cite{robotwin2} benchmark with cluttered object distractions and unseen environments (varying lighting, backgrounds, and heights) as shown in the Table~\ref{ood_result}. Our method consistently outperformed the baseline, validating its effectiveness on OOD tasks. Furthermore, in real-world environments where redundant information increases, our method still maintains good performance, particularly as our method can effectively enable the model to focus on specific interactive objects rather than distracted objects or background as shown in Figure~\ref{process}(b).

\begin{table}[h]
\setlength{\tabcolsep}{0.6pt}
\centering
\caption{Comparison in real-world environments with success rate and inference speed. Our method demonstrates substantial performance improvements over baseline approaches.}
\label{real_world_result}
\renewcommand{\arraystretch}{1.2}
\begin{tabular}{c | cccccc | c}
\hline
\textbf{Method} & \begin{tabular}{@{}c@{}}Bottle \\ Handover\end{tabular} & \begin{tabular}{@{}c@{}}Grasp \\ Chicken\end{tabular} & \begin{tabular}{@{}c@{}}Slide \\ Bag\end{tabular} & \begin{tabular}{@{}c@{}} Insert \\ Cable\end{tabular} & \begin{tabular}{@{}c@{}}Banana \\ Hard\end{tabular} & \textbf{Mean} &\textbf{FPS}  \\
\hline
$\pi_0$~\cite{pi0}       & 0.30 & 0.45 & 0.15 & 0.10& 0.35 & 0.27 & 5.8 Hz\\
$\pi_0$+DART~\cite{DART} & 0.20 & 0.50 & 0.05 & 0.00 & 0.20& 0.19 & 8.5 Hz \\
$\pi_0$+BFA~\cite{bfa}   & 0.45 & \textbf{0.65} & 0.20 & 0.20& 0.40 & 0.38& 10.1 Hz \\
$\pi_0$+BFA++            & \textbf{0.60} & \textbf{0.65} & \textbf{0.30} & \textbf{0.30}& \textbf{0.55} &  \textbf{0.48} & 9.4 Hz\\
\hline
RDT~\cite{rdt} & 0.35 & 0.50 & 0.05  & 0.05& 0.20 & 0.23 &1.2 Hz \\
RDT+DART~\cite{DART} & 0.25 & 0.35  & 0.00 & 0.00 & 0.20& 0.16 &1.8 Hz\\
RDT+BFA~\cite{bfa}   & 0.40 & 0.65  & 0.20 & 0.15 & 0.35& 0.35 &1.8 Hz\\
RDT+BFA++            & \textbf{0.55 }& \textbf{0.75} & \textbf{0.35} &  \textbf{0.20}  & \textbf{0.45} & \textbf{0.46} &1.7 Hz\\
\hline
\end{tabular}
\vspace{-0.3cm}
\end{table}

\begin{table}[h]
\setlength{\tabcolsep}{1.0 pt}
\centering
\caption{Comparison in OOD environments with success rate and inference speed in RoboTwin2. Our method shows impressive improvement.}
\label{ood_result}
\renewcommand{\arraystretch}{1.2}

\begin{tabular}{c | ccccc | c}
\hline
\textbf{Method} & \begin{tabular}{@{}c@{}}Adjust \\ Bottle\end{tabular} & \begin{tabular}{@{}c@{}}Beat \\ Hammer\end{tabular} & \begin{tabular}{@{}c@{}}Handover \\ Block\end{tabular} & \begin{tabular}{@{}c@{}}Open \\ Laptop\end{tabular}  & \textbf{Mean} & \textbf{FPS}  \\
\hline
$\pi_0$~\cite{pi0}       & 0.56 & 0.21 & 0.08 & 0.46 & 0.328 & 6.4 Hz\\
$\pi_0$+DART~\cite{DART} & 0.40 & 0.08 & 0.05 & 0.30 & 0.208 & 10.0 Hz \\
$\pi_0$+BFA~\cite{bfa}   & 0.59 & 0.24 & 0.18 & 0.46 & 0.368 & 11.3 Hz \\
$\pi_0$+BFA++            & \textbf{0.63} & \textbf{0.25} & \textbf{0.19} & \textbf{0.55} & \textbf{0.405} & 10.5 Hz\\
\hline
\end{tabular}

\vspace{-0.8cm}
\end{table}

\subsection{Analysis}
\label{sec:analysis}
We analyze the tokens from both BFA++ and the baseline to understand why pruning actually improves performance.

\begin{figure}[h]
  \centering
    \begin{tabular}{cc}      \includegraphics[width=0.45\linewidth]{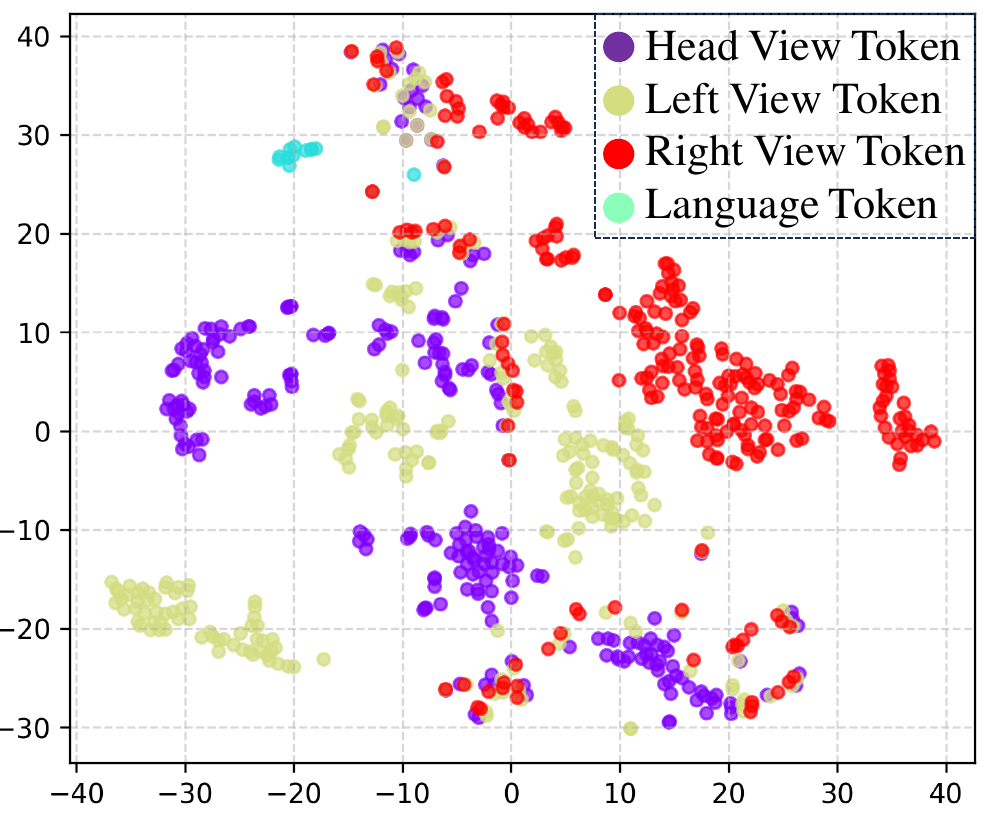}   &  \includegraphics[width=0.45\linewidth]{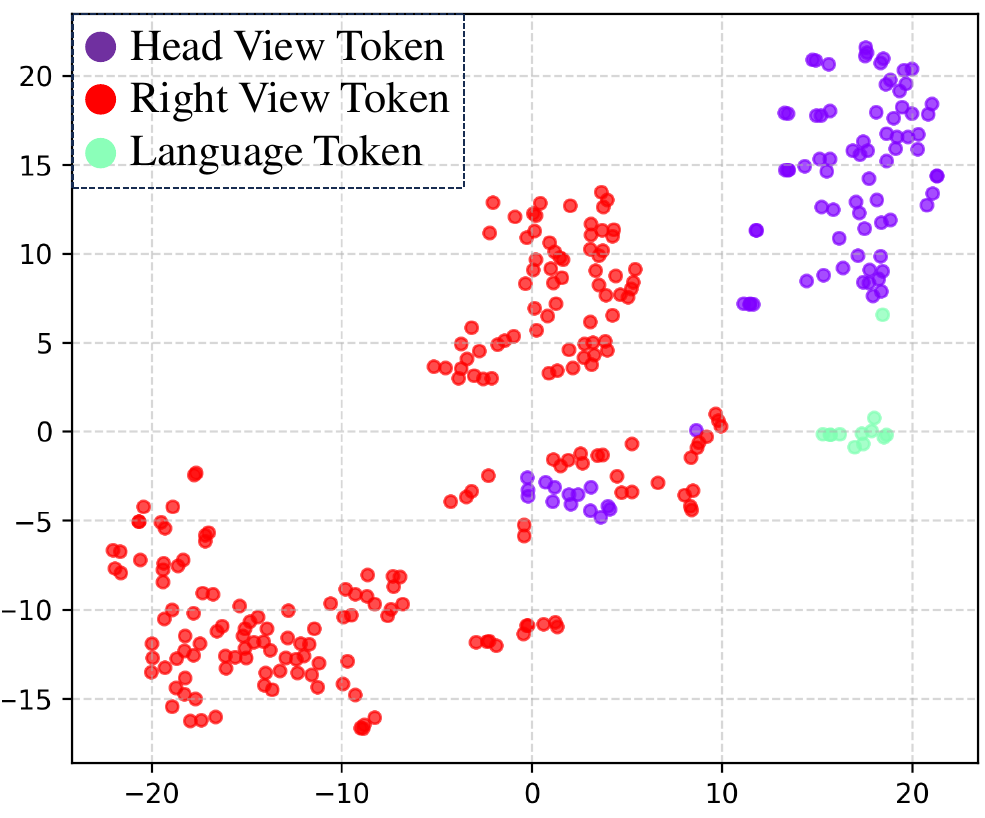}\\
      (a) Baseline~\cite{pi0}   &  (b) Our BFA++
    \end{tabular}

   \caption{T-SNE of token with baseline and our BFA++. Tokens from the left view are omitted as they are entirely pruned by BFA++, and BFA++ shows less similar token distributions across views.}
   \label{point_T_SNE}
   \vspace{-0.3cm}
\end{figure}

First, as shown in Figure~\ref{point_T_SNE}, we perform t-SNE~\cite{T-SNE} visualization on the tokens generated by the baseline and our method on $\pi_0$~\cite{pi0}. We can see that the original tokens from the three views are highly mixed, indicating that the model is processing a lot of redundant information. After BFA++ pruning, tokens from different views are well separated and highly distinct. This shows that input redundancy is almost eliminated, allowing the model to better extract useful information.

\begin{figure}[h]
  \centering
\begin{tabular}{ccc}    \includegraphics[width=0.3\linewidth]{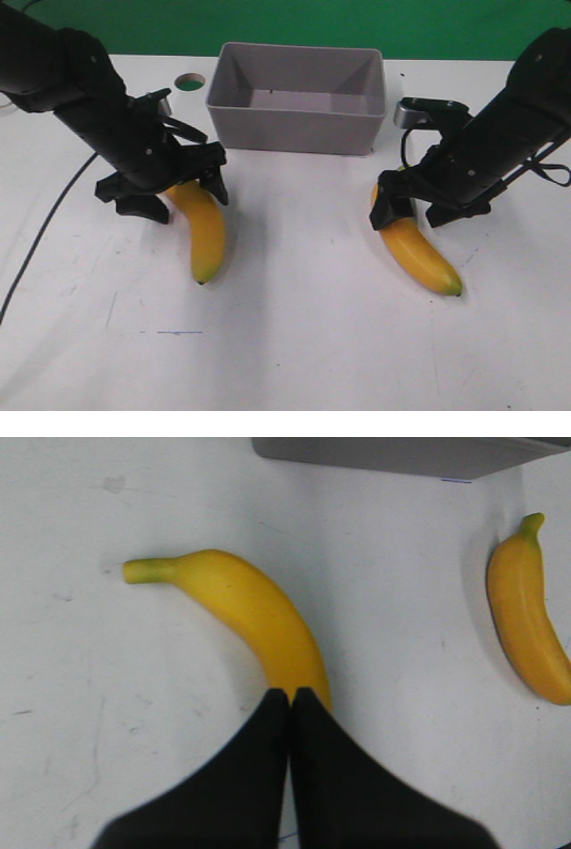} & \includegraphics[width=0.3\linewidth]{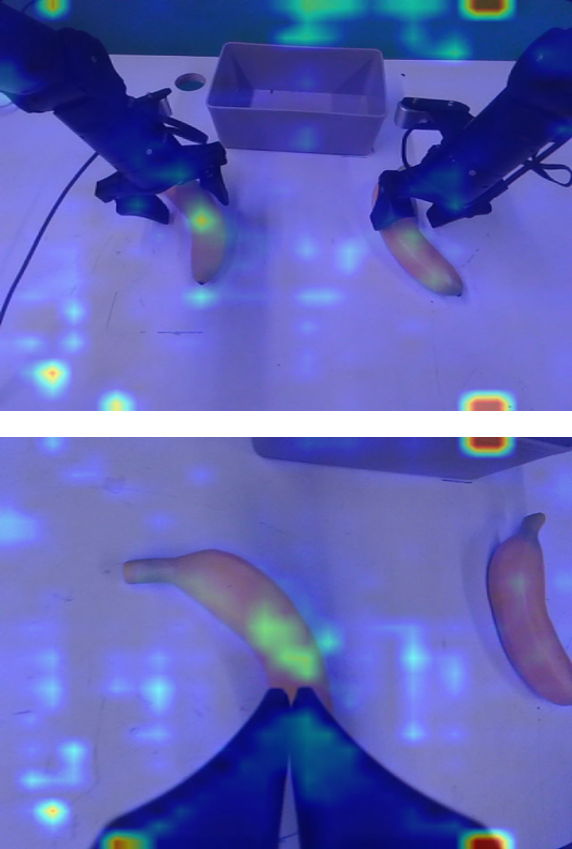} & \includegraphics[width=0.3\linewidth]{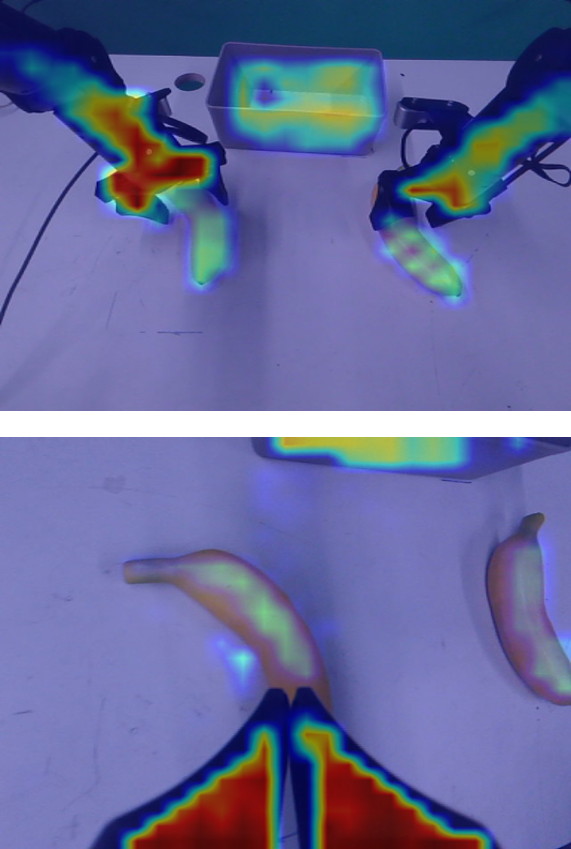}\\
    (a) Input image & (b) Baseline & (c) Our BFA++
\end{tabular}
    
   \caption{Grad-Cam heatmaps with baseline and our BFA++. Our method exhibits significantly more concentrated attention on the gripper and interactive objects.}
   \label{Grad_cam}
   \vspace{-0.5cm}
\end{figure}

Next, we visualize the Grad-CAM \cite{GradCam} heatmaps produced by the vision encoder for both the baseline ($\pi_0$) and our method ($\pi_0$ + BFA++).
As shown in Figure~\ref{Grad_cam}, We find that the baseline model shows scattered attention, focusing on global information. After applying our token pruning method, the vision encoder can focus more on important areas, such as the gripper and interacting objects.
Unlike typical VLMs, global information in VLA scenarios often includes a lot of irrelevant content. Our method helps the model focus its attention on key interaction regions with more useful information.

\subsection{Ablation Study}
We conduct ablation experiments on our algorithm components using $\pi_0$~\cite{pi0} on four simulation tasks.

\begin{table}[h]
\caption{Ablation on components and method of BFA++ with $\pi_0$. 
The results show that all our components are effective and the designed BFA++ pruning structure is superior to alternative methods.}
\centering
\setlength{\tabcolsep}{3pt}

\begin{tabular}{c | c c c c c|c}
\specialrule{1pt}{0pt}{1pt}
\textbf{Method} & \begin{tabular}{@{}c@{}}Block \\ Hammer\end{tabular} & \begin{tabular}{@{}c@{}}Bottle \\ Adjust\end{tabular} & \begin{tabular}{@{}c@{}}Shoe \\ Place\end{tabular} & \begin{tabular}{@{}c@{}}Cup \\ Place\end{tabular} & \textbf{Mean} & \textbf{FPS}\\ 
\specialrule{1pt}{0pt}{1pt}
w/o Inter-IP          & 0.76 & 0.41 & 0.56 & 0.50 & 0.558  & 10.8 Hz\\
w/o Intra-IP         & 0.74 & 0.51 & 0.56 & 0.35 & 0.540  & 11.0 Hz\\
w/o Hierarchical & 0.73 & 0.49 & 0.58 & 0.46 & 0.565  & 10.4 Hz\\
w/o Adaptive Weight & 0.84 & 0.53 & 0.60 & 0.46 & 0.608  & 10.4 Hz\\
\specialrule{1pt}{0pt}{1pt}
Adaptive Pooling & 0.75 & 0.48 & 0.55 & 0.46 & 0.560 & 11.9 Hz \\
Random Drop & 0.70 & 0.51 & 0.52 & 0.41 & 0.535  & 11.5 Hz \\
Online Drop & 0.66 & 0.45 &  0.50 & 0.39 & 0.500 & 1.3 Hz \\
Adaptive Ratio Drop &  \textbf{0.84}&  \textbf{0.60} &  0.63 &  0.46 &  0.633 &  8.4 Hz \\

Our BFA++            & \textbf{0.84} & 0.58 & \textbf{0.65} & \textbf{0.48} & \textbf{0.638} & 10.3 Hz\\
\specialrule{1pt}{0pt}{1pt}
\end{tabular}
\label{ablation_main}
\vspace{-0.3cm}
\end{table}
In Table~\ref{ablation_main}, ``w/o Hierarchical" means we obtain the final token scores by directly multiplying inter-view importance with intra-view importance, and conduct pruning accordingly. Without hierarchical token pruning, our success rate drops significantly. This is because direct end-to-end ranking tends to select the main view while discarding wrist camera information, similar to using only the main view. 
``w/o Adaptive Weight" refers to distance weighting not being used. The results show that using distance weighting, which prevents pruning the tokens between objects and grippers, improves our method's performance in all four tasks. 
Therefore, our hierarchical token pruning and adaptive weight are both effective. 
    

Then we conduct experiments without the inter-view importance predictor (Inter-IP) and intra-view importance predictor (Intra-IP), shown as ``w/o Inter-IP" and ``w/o Intra-IP" in Table~\ref{ablation_main}, using only hierarchical pruning without cross-view importance and only using the inter-view importance score to prune. As shown in Table~\ref{ablation_main}, both components enable our model to more accurately identify which tokens to prune. 
To further validate our method, we compare BFA++ with adaptive pooling~\cite{Deco}, random token pruning, and online pooling, which computes intra-view and inter-view importance scores via online annotation. For fair comparison, all methods are applied during both post-training and inference. Our method significantly outperforms all baselines. We observe that online pruning shows notably lower speed and success rate than the baseline, because online annotation is excessively time-consuming during inference and substantially increases post-training time due to repeated annotations. In contrast, our method effectively backpropagates gradients from both inter-view and intra-view perspectives. Finally, we ablate the fixed prune ratio by comparing with adaptive token prune ratios.
During two-step pruning, it prunes the number of tokens with scores below 0.5 multiplied by 0.8, which is the best parameter we found through exploration, with other settings identical to BFA++. This approach shows inconsistent speed due to variable prune ratio, making it unsuitable for VLA tasks, and achieves lower success rates than BFA++.

\begin{figure}[h]
  \centering
    \includegraphics[width=\linewidth]{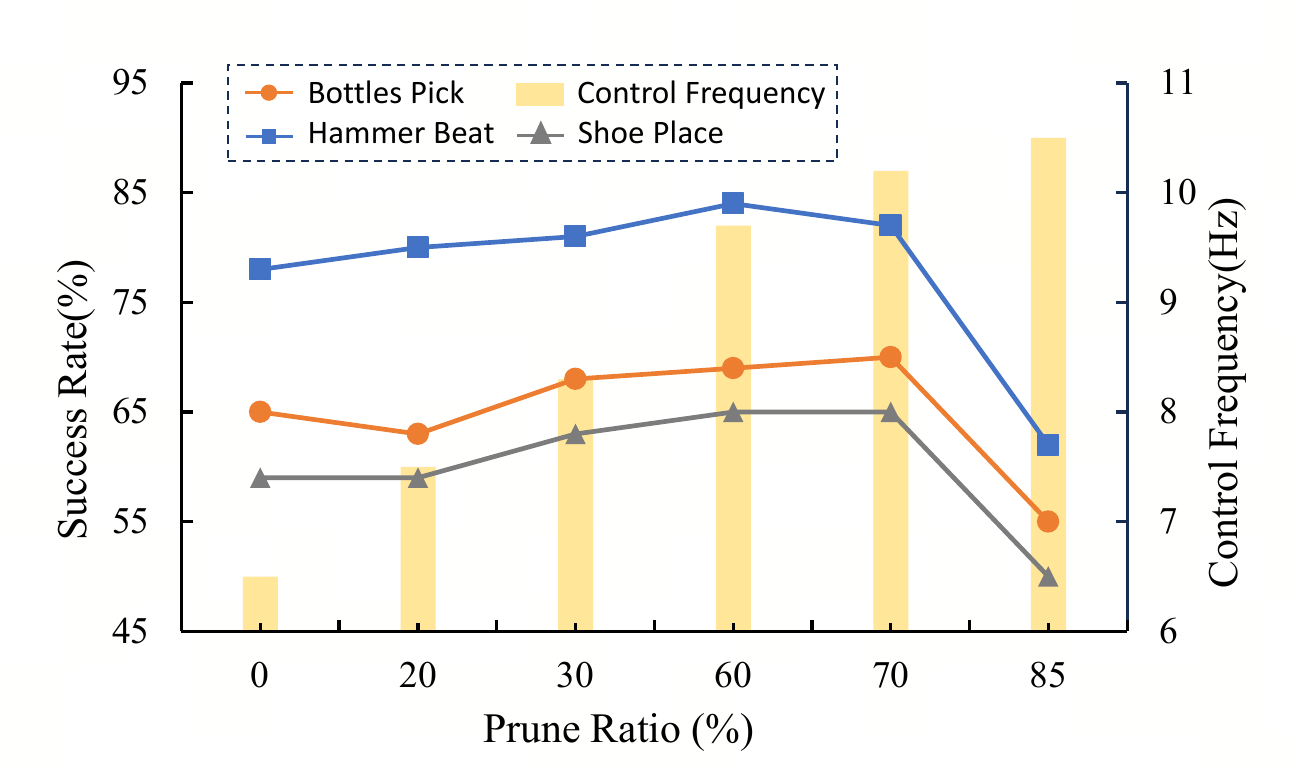}
    
   \caption{The ablation of the final prune ratio of BFA++. As the prune ratio increases, the control frequency accelerates, and the success rate first increases and then decreases. Our choice of prune ratio is balanced, as it not only accelerates the process but also maintains a relatively high success rate.}
   \label{Abl_num}
   \vspace{-0.5cm}
\end{figure}

Next, we conducted an ablation study on the total amount of token pruning by scaling the drop ratio proportionally to assess its effect. As shown in Figure~\ref{Abl_num}, this behavior can be neatly explained by mutual information theory: when the pruning rate is low, both model accuracy and inference speed improve together, because what we remove are mostly redundant or minor tokens that carry little mutual information with the task goal. Dropping them not only reduces computation but also acts as a denoiser, helping the model focus on the truly important bits. However, once the pruning rate exceeds a certain threshold, even as speed continues to improve, accuracy drops sharply. The reason is that pruning begins to remove key tokens that have very high mutual information with the correct answer, so performance degrades.


\section{Conclusion} 
In this work, we propose BFA++, a dynamic token pruning framework that addresses redundant visual information in multi-view vision–language–action models. Our method introduces two-level importance predictors to identify critical camera views and task-relevant regions, jointly trained with the VLA model using automatically annotated data. Through hierarchical token pruning, BFA++ removes redundant tokens while preserving essential manipulation details. Experiments on the RoboTwin benchmark~\cite{robotwin} show that BFA++ achieves 1.5–1.8$\times$ speedup while improving success rates by about 10\% on both $\pi_0$~\cite{pi0} and RDT~\cite{rdt}, demonstrating that intelligent feature selection can outperform brute-force processing of all inputs. As a post-training approach, BFA++ may face generalization challenges under unseen objects or camera configurations, as the importance score networks are optimized for the training distribution. Future work will focus on improving the robustness of the importance predictors to broader variations in objects and viewpoints.

\section*{ACKNOWLEDGMENT}
This work was supported in part by the National Key Research and Development Program of China under Grant 2024YFC3308500, the National Science Foundation of China under Grants 92570206 and 62473356, the Beijing Natural Science Foundation under Grants L232028 and L242060, and the Chongqing Industry and Information Technology R\&D Program under Grant YJX-2025001001003.

\bibliographystyle{IEEEtran}
\bibliography{reference.bib}

\end{document}